\documentclass[conference]{IEEEtran}
\usepackage{cite}
\ifCLASSINFOpdf
  \usepackage[pdftex]{graphicx}
\else

\usepackage[dvipdfmx]{graphicx}

\fi
\graphicspath{{./fig/}}
\usepackage{amsmath}
 \usepackage[caption=false,font=footnotesize]{subfig}

\begin{document}
\title{A Generative Model of Textures Using Hierarchical Probabilistic Principal Component Analysis}

\author{\IEEEauthorblockN{
Aiga SUZUKI\IEEEauthorrefmark{1}\IEEEauthorrefmark{2},
Hayaru SHOUNO\IEEEauthorrefmark{3}
}
\IEEEauthorblockA{
  \IEEEauthorrefmark{1} University of Tsukuba\\
  1-1-1 Tennodai, Tsukuba, Ibaraki, 305-0006, Japan\\
}

\IEEEauthorblockA{
  \IEEEauthorrefmark{2} National Institute of Advanced Industrial Science and Technology\\
  1-1-1 Umezono, Tsukuba,  Ibaraki, 305-0045, Japan\\
  Email: ai-suzuki@aist.go.jp
}

\IEEEauthorblockA{
  \IEEEauthorrefmark{3} University of Electro-Communication\\
  1-5-1 Chofugaoka, Chofu, Tokyo, 182-8585, Japan\\
  Email: shouno@uec.ac.jp 
}

}


%

\IEEEspecialpapernotice{Preprint of a published conference proceeding in PDPTA'17. Please refer to:\\
Aiga Suzuki, Hayaru Shouno,
``Generative Model of Textures Using Hierarchical Probabilistic Principal Component Analysis'', 
Proc. of the 2017 Intl. Conference on Parallel and Distributed Processing Techniques and Applications (PDPTA’17), CSREA Press, pp. 333-338 (2017)
}

\maketitle

\begin{abstract}
Modeling of textures in natural images is an important task to make a microscopic
model of natural images.
Portilla and Simoncelli proposed a generative texture model,
which is based on the mechanism of visual systems in brains,
with a set of texture features and a feature matching.
On the other hand, the texture features, used in Portillas' model, have redundancy between its components
came from typical natural textures.
In this paper, we propose a contracted texture model which provides a dimension reduction for the Portillas' feature.
This model is based on a hierarchical principal components analysis using known group structure of the feature.
In the experiment, we reveal 
effective dimensions to descrive texture is fewer than the original description.
Moreover, we also demonstrate how well the textures can be synthesized 
from the contracted texture representations.
\end{abstract}

\begin{IEEEkeywords}
  texture modeling, thexture synthesis, dimensionality reduction,
  probabilistic principal component analysis
\end{IEEEkeywords}


%
\IEEEpeerreviewmaketitle

\section{Introduction}

From the microscopic point of view,
all of natural scenes, providing our visual stimulus,
can be seen as a patchwork of various texture patterns.
As a biological knowledges, our visual system recognize
the objects by their shapes in primary visual cortex.
In addition to this, recent studies have shown that 
their textures also be important property for recognizing and segmentation
of the objects in our higher-level visual cortex
\cite{adelson2001seeing}.
Thus, availability of texture description is important for
modeling of natural images in the field of computer vision.
Especially, our research focus is to construct a generative
model for various texture patterns.

Most of recent studies of texture modeling,
achieving a good benchmark results,
are based on the Markov Random Field (MRF) and Gibb's sampling
\cite{heess2009learning,Ranzato2010,Luo2012,Kivinen2012,Ranzato2013}.
These MRF based models assume that
textures are characterized as just a 2-dimensional
probabilistic distributions.
Therefore, the most MRF based texture models would have lack of consideration
for ``\textit{a perceptually equivalence}''.

\begin{figure}[t]
\begin{minipage}[b]{0.45\columnwidth}
\centering
\includegraphics[keepaspectratio,width=1.2in]{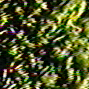}
\end{minipage}
\begin{minipage}[b]{0.45\columnwidth}
\centering
\includegraphics[keepaspectratio,width=1.2in]{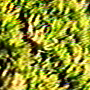}
\end{minipage}
\caption{
The macro photographs of same surfaces but  different light sources
from ``Moss'' class of CUReT texture database\cite{dana1999reflectance}.
(Left) front faced and low light amount. (Right) oblique faced and high light amount.
}
\label{fig:moss_comparison}
\end{figure}

The perceptually equivalence is our perceptual property considering that
``the pattern A and B are came from same texture''.
The perceptually equivalence is not referred explicitly from structural modeling,
such as MRF models, but also naturally required as in a part of
modeling of natural images.
Figure \ref{fig:moss_comparison} shows the example of the perceptually equivalence,
which are the pictures of same materials but different illuminance.
We can find that the structural/geometric differences,
that are overexposures and shadow directions,
came from the different light source, on the other hand,
we can also decide that the both of them are same textures.

Portilla and Simoncelli proposed a texture feature,
which considers such a perceptual aspects,
and a generative model on the basis of feature matching reconstruction
\cite{portilla1996texture,Portilla2000}.
Hereafter, we call their texture feature,
``\textit{a Portilla-Simoncelli statistics}'', PSS for short.
The PSS is influenced by bilogical knowledges,
the receptive field of the primary visual cortex 
and Gabor-like filters especially in the area V1.
The PSS is based on wavelet-like multi-scale image decomposition method,
which has translation-invariance and rotation-invariance,
known as a steerable filter pyramid\cite{simoncelli1995steerable}.
The PSS is consists of a marginal statistics over
decomposed images, which act as constraints of the texture structure.
Simoncelli et al. showed that the PSS could charactarize and generate
various textures
\cite{portilla1996texture,portilla1999texture,simoncelli1998texture,Portilla2000}.
In recent, moreover, some studies report that the PSS could express the
selective neuronal activity in area V4 of macaque \cite{Okazawa2015}
and in area V2 of humans' cerebral cortex \cite{Hiramatsu2011}.
These reports were implying that the PSS could be appropriate representation
of our texture recognition mechanisms.

Considering the modeling of textures in natural images,
hereafter we call it as ``natural textures'',
the PSS would be good representation, however, the PSS would have redundancy between
its elements, because the natural textures typically 
have certain structure.
Therefore, the PSS extracted by natural textures could be
phrased with more simplified representation.

This paper proposes a dimension reduction method to grasp the latent factors in
the PSS of the natural textures.
Our method is based on Probabilistic Principal Component Analysis (PPCA) 
and focusing on known correlation of the PSS.
We achieve a 88.8[\%] dimension reduction from raw PSS
to phrase natural texture dataset, which is hard to apply a plain PCA.

\section{A Portilla-Simoncelli statistics}

This section gives an overview of the PSS used in
our study as a texture feature.

\subsection{A steerable filter pyramid}

A steerable filter pyramid 
\cite{simoncelli1995steerable,portilla1999texture}
is multi-scale and directive
image decomposition method which partially imitates
the orientation selectivity of humans' visual systems.
The steerable filter pyramid could be ``\textit{steered}''
its decomposition traits
by 2 parameters, a number of decomposition scales $N$
and a number of decomposition orientations/directions $K$,
thus, this name was given.
These properties originated in complex orthogonal Wavelet transform
and Gabor filters bank.

\begin{figure*}[t]
\centering
\includegraphics[width=6in]{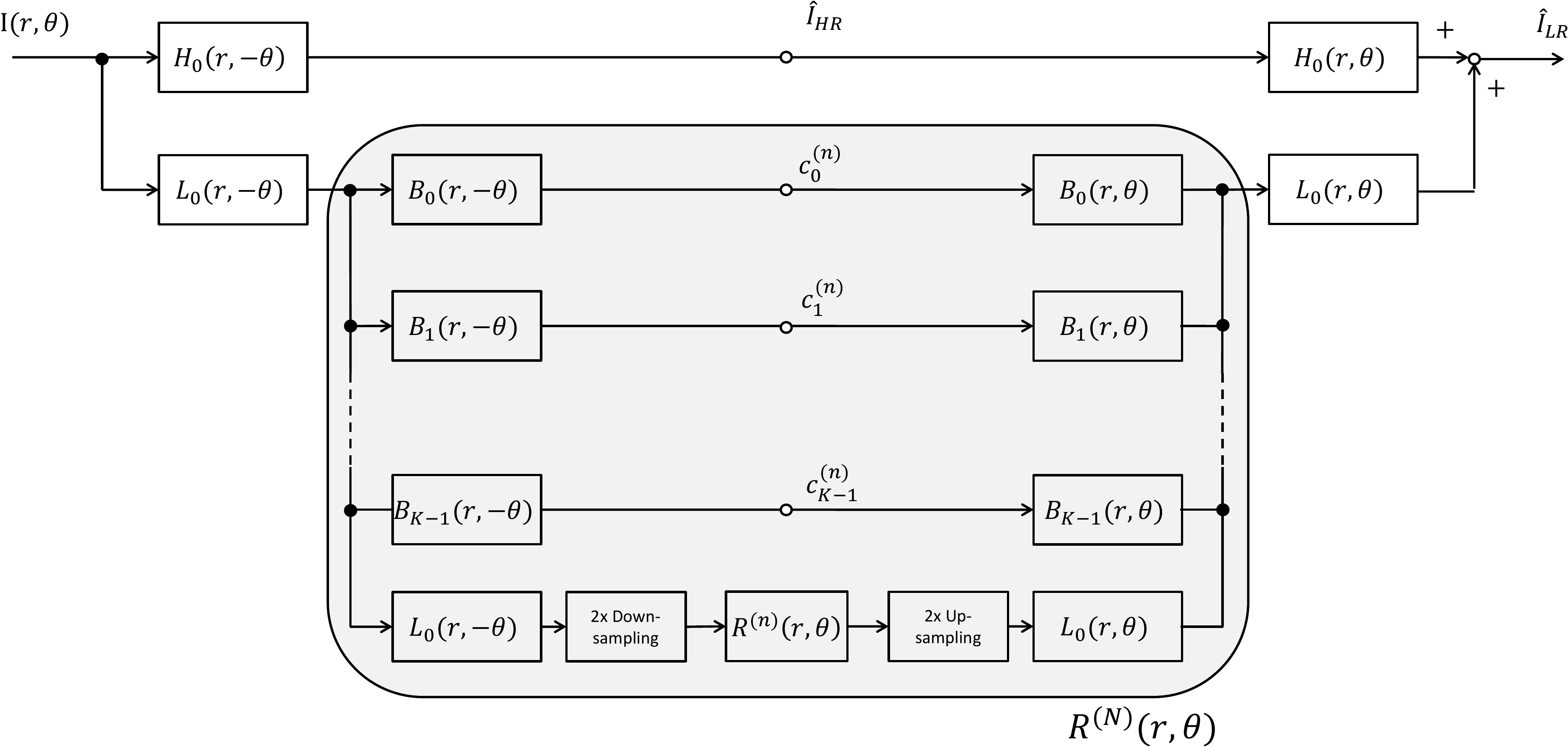}
\caption{
Block diagram of steerable filter pyramid in Fourier domain.
The white circle denotes a observation point.
}
\label{fig:sfp_blk_diagram}
\end{figure*}

The steerable filter pyramid was defined in Fourier domain,
in similar to wavelet transform.
Figure \ref{fig:sfp_blk_diagram}
shows a Fourier domain 
block diagram of a steerable filter pyramid
interpreted as a linear system.
Let $I(x, y)$ be an input image in spatial domain,
and $\tilde{I}(r, \theta)$ be a polar representation of
an input image in Fourier domain,
each transfer elements in Figure \ref{fig:sfp_blk_diagram}
would be given by:

\begin{align}
L(r, \theta) =
\begin{cases}
2 \cos \left(\dfrac{\pi}{2} \log_2\left(\dfrac{4r}{\pi} \right) \right),\qquad &\left(\dfrac{\pi}{4} < r < \dfrac{\pi}{2} \right) \\
2, \qquad & \left(r \leq \dfrac{\pi}{4} \right) \\
0, \qquad & \left(r \geq \dfrac{\pi}{2} \right)
\end{cases}
\label{eq:spf_lowpass}
\end{align}

\begin{align}
B_k(r, \theta) = H(r) G_k(\theta),\quad k\in 0, \dots, K-1,
\end{align}
the band-pass transfer element $B_k(r, \theta)$ was
decomposed into radial part $H(r)$ and angular part $G_k(\theta)$
defined as

\begin{align}
H(r) =
\begin{cases}
\cos \left(\dfrac{\pi}{2} \log_2 \left(\dfrac{2r}{\pi} \right) \right),\qquad &\left(\dfrac{\pi}{4} < r < \dfrac{\pi}{2} \right) \\
1, \qquad & \left(r \leq \dfrac{\pi}{4} \right) \\
0, \qquad & \left(r \geq \dfrac{\pi}{2} \right)
\end{cases}
\end{align}

\begin{align}
G_k(\theta) =
\begin{cases}
\alpha_K \left\{ \cos\left(\theta-\dfrac{\pi k}{K}\right)  \right\}^{K-1},\quad & \left(\left\lvert \theta - \dfrac{\pi k}{K} \right\rvert < \dfrac{\pi}{2}\right)\\
0,\quad & (\text{otherwise})
\end{cases}
\end{align}
$\alpha_k$ denotes a coordinate of $B_k$ which depends on
number of orientation $K$, given by 

\begin{align}
\alpha_K = 2^{K-1} \dfrac{(K-1)!}{\sqrt{K(2(K-1))!}}
\end{align}

Low-pass transfer element $L_0$ had to be defined
to reject a band exeeding the Nyquist limit, 
to make this decomposition a complete system.

\begin{align}
L_0(r, \theta) = \dfrac{L(r/2, \theta)}{2}
\label{eq:lowpass}
\end{align}

Refer to Eq.\ref{eq:lowpass}, the high-pass transfer element $H_0$
follows the low-pass residual as
\begin{align}
H_0(r, \theta) = H\left(\dfrac{r}{2}, \theta\right)
\end{align}

As shown in Figure
\ref{fig:sfp_blk_diagram},
the steerable filter pyramid would consist of
recursion of subsystem $R^{(n)}(r, \theta)$.
The multi-scaleness of the steerable filter pyramid was provided by
the recursion of them,
and the orientation selectivity was provided by a parallelism of them.

\subsection{Construction of the PSS}
The PSS is constructed as a list of 10-types of 
statistics $C_1 \sim C_{10}$, came from the steerable filter pyramid.

\subsubsection{Descriptive statistics of pixel values ($C_1$)}
Mean, variance, skewness, kurtosis, minimum and maximum values of
pixel value of the input image.\\
(6 dimensions)

\subsubsection{Descriptive statistics of each scales ($C_2$)}
Skewness and kurtosis of reconstructed images with each scales of steerable filter pyramid,
including a low-pass residual $\tilde{I}_{LR}$.\\
($2(N+1)$ dimensions)

\subsubsection{Auto-correlation of each decompositions ($C_3$)}
$M$-neighbor auto-correlation of reconstructed images with each decompositions 
(scales and orientations) of the steerable filter pyramid.\\
($N\cdot K\cdot M^2$ dimensions)

\subsubsection{Auto-correlaton of each scales ($C_4$)}
$M$-neighbor auto-correlation of reconstructed images with each scales of steerable filter pyramid,
including a low-pass residual $\tilde{I}_{LR}$.\\
($M^2\cdot (N+1)$ dimensions)

\subsubsection{Cross-correlations between decomposed images for each scales ($C_5$)}
Cross-correlation of decomposed image with each scales between each orientations.\\
($N\cdot K^2$ dimensions)

\subsubsection{Cross-correlations between reconstructed images for each scales ($C_6$)}
Cross-correlation of reconstructed images with each scales,
including a low-pass residual $\tilde{I}_{LR}$,
between each orientations.\\
($K^2(N+1)$ dimensions)

\subsubsection{Cross-correrations between each reconstructed images ($C_7$)}
Cross-correration of reconstructed images with each scales,
including a low-pass residual $\tilde{I}_{LR}$,
and each orientations.\\
($K^2\cdot N(N+1)$ dimensions)

\subsubsection{Cross-correlations between each decomposed images for each decompositions ($C_8$)}
Cross-correlation of decomposed image for each decompositions
(scales and orientations).\\
($N^2 \cdot K^2$ dimensions)

\subsubsection{Means of each reconstruted images ($C_9$)}
Means of pixel values of reconstructed images with each decompositions,
including a low-pass residual $\tilde{I}_{LR}$ and high-pass residual $\tilde{I}_{HR}$.\\
($N\cdot K + 2$ dimensions)

\subsubsection{Variance of high-pass residual ($C_{10}$)}
Variance of pixel values of reconstructed images with high-pass residual $\tilde{I}_{HR}$.\\
(1 dimension)
 
Dimensionality of the PSS will be determined by
parameters of steerable filter pyramid $N$, $K$,
and neighbor of pixel space $M$.
In our experiments, we use parameters as
$N=4, K=4, M=7$ according to \cite{Portilla2000} consistently.
The dimensionality of the PSS will be 1784 by these parameters.

\begin{figure}[t]
\centering
\includegraphics[width=3in]{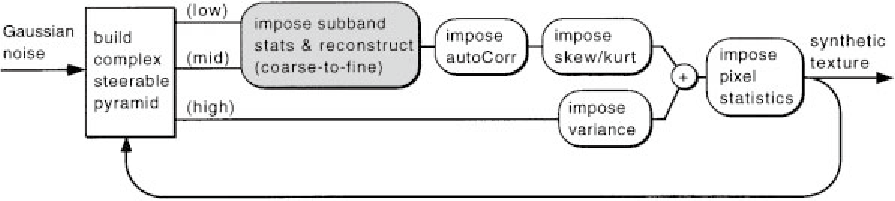}
\caption{
Flowchart of texture reconstruction from the PSS,
quoted from \cite{Portilla2000}.
Texture reconstruction is based on
coarse-to-fine optimization from
the Gaussian white noize.
}
\label{fig:synthesis_diagram}
\end{figure}

Figure \ref{fig:synthesis_diagram} shows a schematic diagram of
texture reconstruction from the PSS.
Texture reconstruction will be accomplished with iterative
optimization from the Gaussian white noize images.
\cite{Portilla2000} shows this reconstruction would converge on desired PSS
with in 50-iterations experimentally.

\section{Hierarchical Probabilistic Principal Component Analysis}

In this paper, we propose a probabilistic principal component analysis (PPCA)
\cite{tipping1999}
based model for dimension reduction, which focuses to known structure of the PSS.
PPCA is a statistical method to estimate the latent variables which
generate an input data essentially.
It is similar to deterministic principal component analysis (PCA) and
provides a same result usually,
however, PPCA is a probabilistic model 
which assume that the input data would be generated under the
Gaussian distribution and the Gaussian noize.
Therefore, PPCA has a merit of an optimization under the high-dimensional
input space.

PPCA could be of use for dimension reduction in most cases,
however, 
we could sometimes know the structure or correlation of the input
data in advance.
For example, construction of the PSS was built up gradually with
10-types of statistics.
In other words, the PSS has a group-structure between its components.
Considering such a known structure,
it is possible to grasp more effective contracted representation of
the input data.

\begin{figure}[t]
\centering
\includegraphics[width=3.5in]{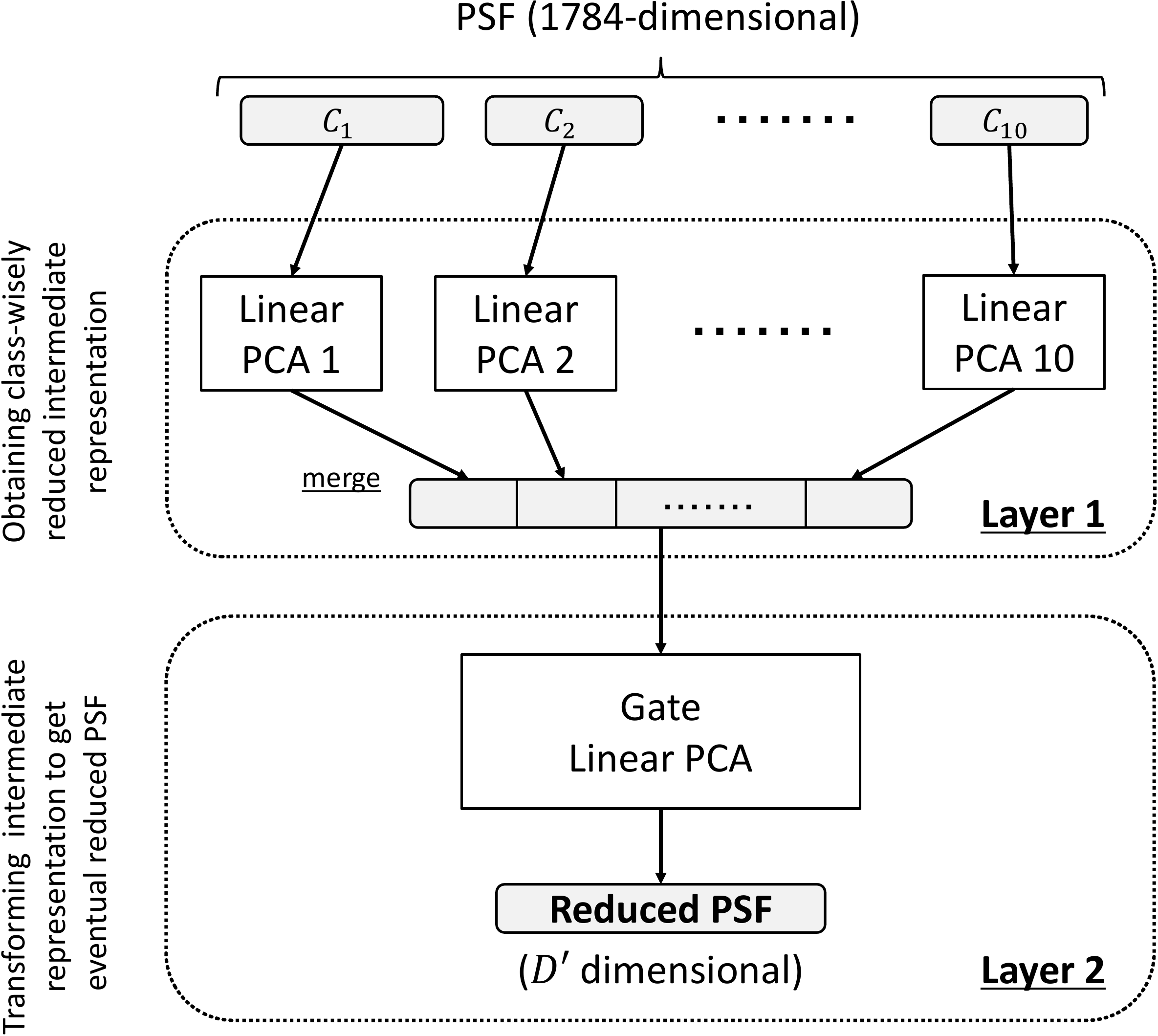}
\caption{
Schematic diagram of the hierarchical probabilistic principal component analysis.
Input will be divided into known groups and contracted with distinct linear-PPCA.
After that, the final linear-PPCA makes a conclusive reduced representation.
}
\label{fig:hppca}
\end{figure}

We propose a novel architecture of PPCA, considering such a known
structure of the input data, a ``hierarchical probabilistic principal
component analysis'' (HPPCA).
Figure \ref{fig:hppca} shows a schematic diagram of the HPPCA.
The HPPCA applies a hierarchical dimension reduction which is based on
a structure of the PSS.
First, the HPPCA applies a dimension reduction for each classes 
$C_1\sim C_{10}$ with distinct PPCA models.
Second, reduced representations will be concatenated into an intermediate
vector, finally, conclusive reduced representation will be given
as an output of the final PPCA.

\section{Materials}

\subsection{Texture Dataset}

\begin{figure}[t]
\centering
\includegraphics[width=2.5in]{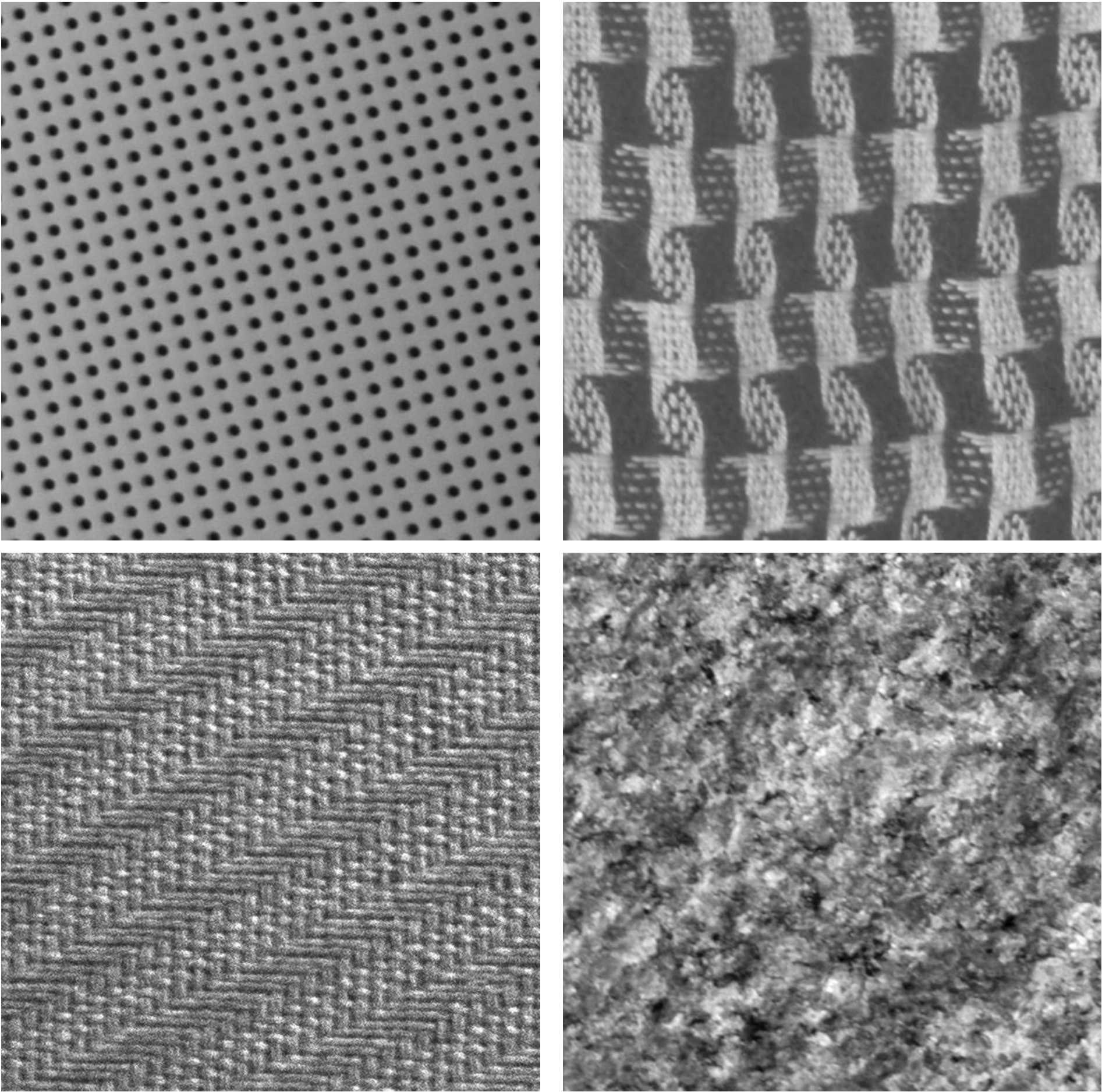}
\caption{
Examples of texture images in Kylberg Texture Dataset.
(top-left) the metal plate ceiling. (top-right) the woven scarf.
(bottom-left) the woven fabric on chair. (bottom-right) Flat part of a granite.
}
\label{fig:kylberg}
\end{figure}

We evaluated our model on natural textures
from Kylberg Texture Dataset v. 1.0
\cite{Kylberg2011c}.
Figure \ref{fig:kylberg} shows examples of texture images in
Kylberg Texture Dataset.
This dataset contains 28 classes of natural textures, which are the
macro photographs of real-world surfaces.
Each classes have 1920 patches of gray-scale images
normalized with a mean value of 127 and a 
standard deviation of 40.
The patches have a resolution of 576$\times$576 pixels
and resized into 128$\times$128 pixel to adjust
the model input.
To evaluate the model performance, we used 1720 patches for 
training, and 200 patches for evaluation.

\subsection{Performance index of texture similarity}

To evaluate the performance of texture reconstruction,
we chose a Texture Similarity Score (TSS) as an
index of texture simirality 
\cite{heess2009learning}.
For a source texture image $\mathbf{x}$ and generated texture sample
$\mathbf{s}$, the TSS will be given by:

\begin{align}
\text{TSS}(\mathbf{s}, \mathbf{x}) := \text{max}\left\{\dfrac{\mathbf{x}_{(1)}^T \mathbf{s}}{\Vert\mathbf{x}_{(1)}\Vert \Vert\mathbf{s}\Vert},\dots,\dfrac{\mathbf{x}_{(\mathcal{I})}^T \mathbf{s}}{\Vert\mathbf{x}_{(\mathcal{I})}\Vert \Vert\mathbf{s}\Vert}\right\}
\label{eq:TSS}
\end{align}
$\mathbf{x}_{(i)}$ denotes patch $i$ within the test region of the image
and $\mathcal{I}$ is the number of possible unique patches in the test region.
In other words, the TSS denotes the maximum cosine similarity between sample patches
and possible source texture region, known as ``the maximum normalized cross correlation''.

A patch and sample of size $19 \times 19$ pixel was adopted to define the TSS
in our experiments, according to previous work \cite{heess2009learning}.


\section{Experiment}

\subsection{A preliminary experiment: dimension reduction with conventional PPCA}

\begin{figure}[t]
\begin{minipage}[b]{0.45\columnwidth}
\centering
\includegraphics[keepaspectratio,width=1.2in]{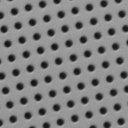}
\end{minipage}
\begin{minipage}[b]{0.45\columnwidth}
\centering
\includegraphics[keepaspectratio,width=1.2in]{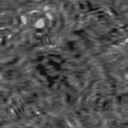}
\end{minipage}
\caption{
Result of reconstruction of the texture image from ceiling1 with the plain linear-PPCA in failure.
(left) source texture. (right) synthesized texture from 1000-dimensional representation.
}
\label{fig:plain_ppca}
\end{figure}

As a preliminary experiment, we tried to obrain a dimension-reduced representation
with plain linear-PPCA directly.
Figure \ref{fig:plain_ppca} shows a reconstruction result by
1784 to 1000 dimension reduction.
The reconstruction result does not reproduce the source texture structure
by the little dimension reduction which expect to preserve the
source PSS structure.
This result implies that the plain PPCA could not grasp the latent
variables of the PSS sufficiently.

\subsection{Appropriate dimensionality of HPPCA}

Dimension reduction with the HPPCA would be steered by 2 part,
to get an intermediate representation and to obtain a conclusive
representation, respectively.
Because of the top-down dimension reduction,
we have to determine a dimension reduction rate of
intermediate PPCAs at first.
We chose the cumulative contribution ratio 
as an index of dimension reduction of each PPCAs.
We determined the dimensions of each PPCAs by variation
of the TSS with cumulative contribution ratio.

\begin{figure}[t]
\centering
\includegraphics[width=3in]{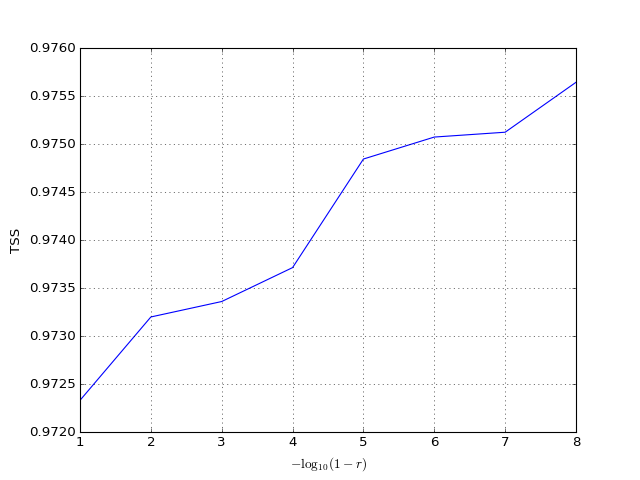}
\caption{
Variation of the TSS by intermediate cumulative contribution ratio.
}
\label{fig:engine_tss_graph}
\end{figure}

Plot of the TSS vs cumulative contribution ratio of intermediate PPCAs
is shown in Figure \ref{fig:engine_tss_graph}.
It was shown that the TSS was monotonically increased as the cumulative
contribution ratio,
however, it exeeded machine epsion of 32-bit float
after $\log_{10}(1-r)=-8,\ (r=0.99999999)$.
Thus, we chose $r=0.99999999$ and 965-dimensional intermediate representation.

\begin{figure}[t]
\centering
\includegraphics[width=3in]{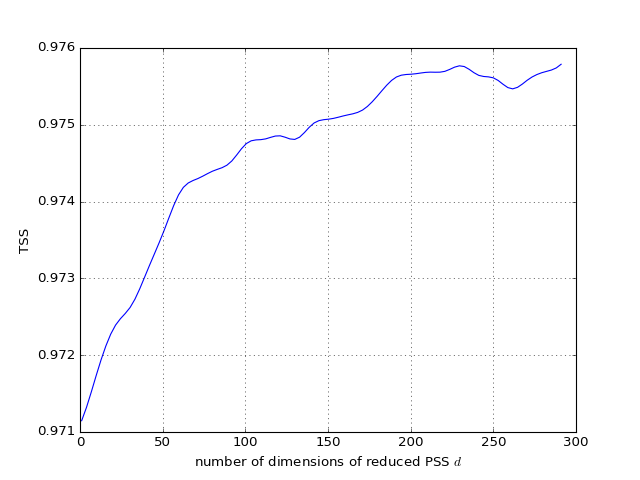}
\caption{
Variation of the TSS by dimensionality of model outputs.
}
\label{fig:gate_tss_graph}
\end{figure}

By given intermediate layer,
we chose a conclusive dimensionality of model output.
Plot of the TSS vs conclusive dimensionality of the HPPCA
is shown in Figure \ref{fig:gate_tss_graph}.
The TSS seemed to saturate around $d$ equal 150 to 200.
We adopt $d=200$ as conclusive reduced representation of
PSS in our experiments.

\subsection{Qualitative evaluation}
\begin{figure*}[t]
\centering
\includegraphics[width=6in]{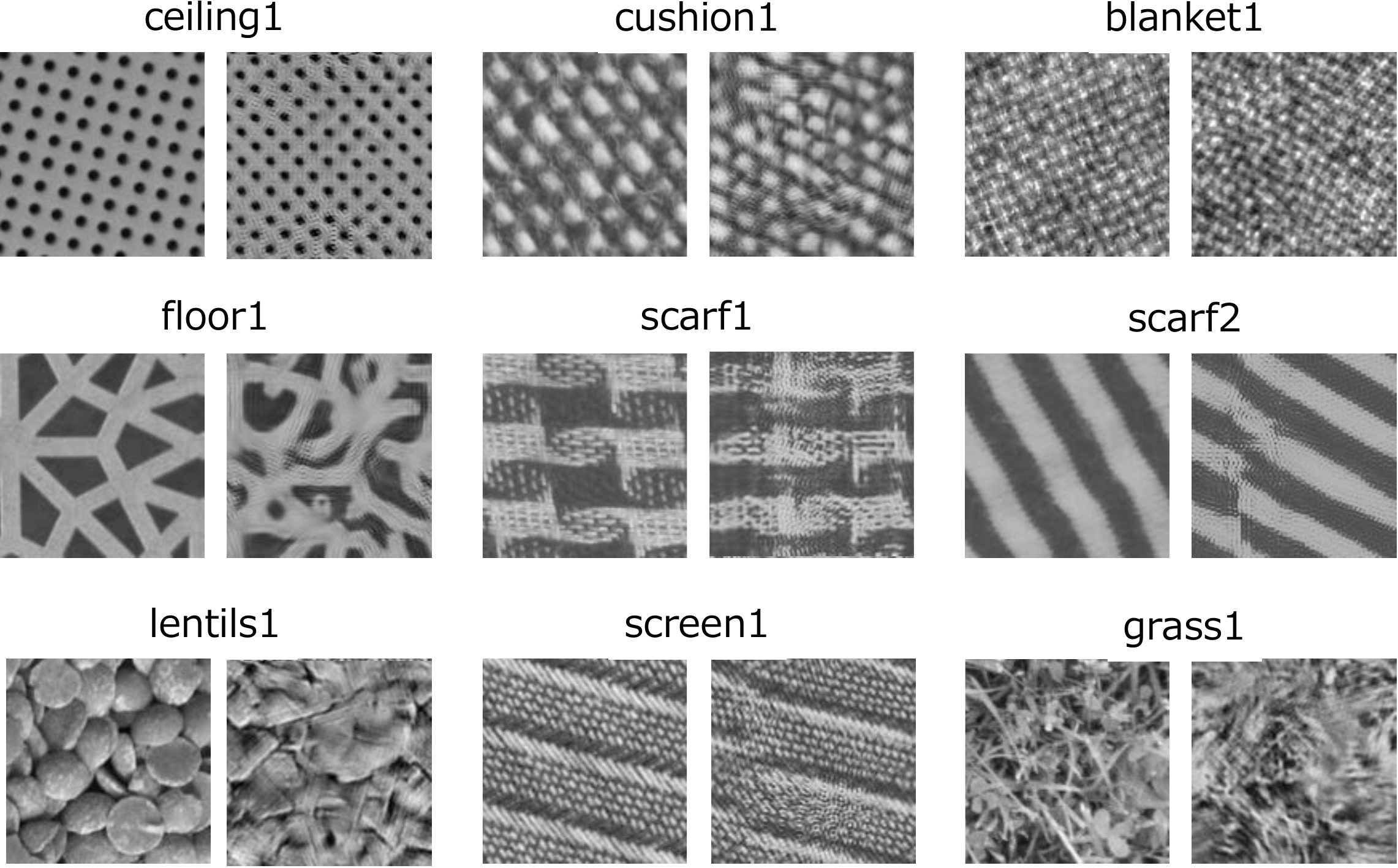}
\caption{
Reconstruction results from 200-dimensional contracted TSS.
The left side of each classes denotes the source texture,
right side denotes the reconstructed texture.
}
\label{fig:recon}
\end{figure*}

This section shows that the contracted representation by our proposed method
can reconstruct the textures with perceptually equivalence.
The reconstructed textures by contracted PSS using the HPPCA disscussed
in previous section shown in Figure \ref{fig:recon}.
The fine textures, that are ceiling1, cushion1, and blanket1, were
well-reproduced by reduced PSS (\ref{fig:recon} first row).
On the other hand, however, the coarse textures, which have large patterns, 
sometimes have trouble with a reproducing the continuous
(\ref{fig:recon} second and third row).
This could be due to the lack of reproduce the low-frequency component
of reduced PSS with the HPPCA.
Nevertheless, the most of reconstruction results might be said to
reproduce source texture structure and we could determine
these textures were almost much the same.

\section{Discussion and conclusion}

The dimension reduction method of the Portilla-Simoncelli statistics,
that is a perceptual texture feature, using the hierarchical probabilistic
principal component analysis was introduced in this paper.
We achieved a 88.8 [\%] dimension reduction from the raw PSS preserving the
source texture structures in reconstruction.

The HPPCA model could be read as the Gaussian-Gaussian restricted
Boltzmann machine adopted a sparsity in its connections \cite{karakida2016dynamical}.
Such connectionism would be make the PPCA easier to grasp the
latent structure of the input space.
In future work, we intend to analyze mathematically the machanizm of
the HPPCA model and build more generalized texture model
via the large-scaled natural image datasets. 

\bibliographystyle{IEEEtran}
%
\bibliography{bib}

\begin{thebibliography}{10}
\providecommand{\url}[1]{#1}
\csname url@samestyle\endcsname
\providecommand{\newblock}{\relax}
\providecommand{\bibinfo}[2]{#2}
\providecommand{\BIBentrySTDinterwordspacing}{\spaceskip=0pt\relax}
\providecommand{\BIBentryALTinterwordstretchfactor}{4}
\providecommand{\BIBentryALTinterwordspacing}{\spaceskip=\fontdimen2\font plus
\BIBentryALTinterwordstretchfactor\fontdimen3\font minus
  \fontdimen4\font\relax}
\providecommand{\BIBforeignlanguage}[2]{{%
\expandafter\ifx\csname l@#1\endcsname\relax
\typeout{** WARNING: IEEEtran.bst: No hyphenation pattern has been}%
\typeout{** loaded for the language `#1'. Using the pattern for}%
\typeout{** the default language instead.}%
\else
\language=\csname l@#1\endcsname
\fi
#2}}
\providecommand{\BIBdecl}{\relax}
\BIBdecl

\bibitem{adelson2001seeing}
E.~H. Adelson, ``On seeing stuff: the perception of materials by humans and
  machines,'' in \emph{Photonics West 2001-electronic imaging}.\hskip 1em plus
  0.5em minus 0.4em\relax International Society for Optics and Photonics, 2001,
  pp. 1--12.

\bibitem{heess2009learning}
N.~Heess, C.~K. Williams, and G.~E. Hinton, ``Learning generative texture
  models with extended fields-of-experts.'' in \emph{BMVC}, 2009, pp. 1--11.

\bibitem{Ranzato2010}
M.~Ranzato, V.~Mnih, and G.~Hinton, ``{Generating more realistic images using
  gated MRF's},'' \emph{Nips}, pp. 1--9, 2010.

\bibitem{Luo2012}
\BIBentryALTinterwordspacing
H.~Luo, P.~L. Carrier, A.~Courville, and Y.~Bengio, ``{Texture Modeling with
  Convolutional Spike-and-Slab RBMs and Deep Extensions},'' \emph{Proceedings
  of the 16th International Conference on Artificial Intelligence and
  Statistics (AISTATS)}, vol.~31, pp. 415--423, 2012. [Online]. Available:
  \url{http://arxiv.org/abs/1211.5687}
\BIBentrySTDinterwordspacing

\bibitem{Kivinen2012}
\BIBentryALTinterwordspacing
J.~J. Kivinen and C.~K.~I. Williams, ``{Multiple Texture Boltzmann Machines},''
  \emph{Proceedings of the 15th International Conference on Artificial
  Intelligence and Statistics (AISTATS 2012)}, vol.~22, pp. 638--646, 2012.
  [Online]. Available:
  \url{http://homepages.inf.ed.ac.uk/s0960152/papers/MTBM-AISTATS12.pdf}
\BIBentrySTDinterwordspacing

\bibitem{Ranzato2013}
M.~Ranzato, V.~Mnih, J.~M. Susskind, and G.~E. Hinton, ``{Modeling natural
  images using gated MRFs},'' \emph{IEEE Transactions on Pattern Analysis and
  Machine Intelligence}, vol.~35, no.~9, pp. 2206--2222, 2013.

\bibitem{dana1999reflectance}
K.~J. Dana, B.~Van~Ginneken, S.~K. Nayar, and J.~J. Koenderink, ``Reflectance
  and texture of real-world surfaces,'' \emph{ACM Transactions on Graphics
  (TOG)}, vol.~18, no.~1, pp. 1--34, 1999.

\bibitem{portilla1996texture}
J.~Portilla, R.~Navarro, O.~Nestares, and A.~Tabernero, ``Texture
  synthesis-by-analysis method based on a multiscale early-vision model,''
  \emph{Optical Engineering}, vol.~35, no.~8, pp. 2403--2417, 1996.

\bibitem{Portilla2000}
J.~Portilla and E.~P. Simoncelli, ``{Parametric texture model based on joint
  statistics of complex wavelet coefficients},'' \emph{International Journal of
  Computer Vision}, vol.~40, no.~1, pp. 49--71, 2000.

\bibitem{simoncelli1995steerable}
E.~P. Simoncelli and W.~T. Freeman, ``The steerable pyramid: A flexible
  architecture for multi-scale derivative computation,'' in \emph{Image
  Processing, 1995. Proceedings., International Conference on}, vol.~3.\hskip
  1em plus 0.5em minus 0.4em\relax IEEE, 1995, pp. 444--447.

\bibitem{portilla1999texture}
J.~Portilla and E.~P. Simoncelli, ``Texture modeling and synthesis using joint
  statistics of complex wavelet coefficients,'' in \emph{IEEE workshop on
  statistical and computational theories of vision}, 1999.

\bibitem{simoncelli1998texture}
E.~P. Simoncelli and J.~Portilla, ``Texture characterization via joint
  statistics of wavelet coefficient magnitudes,'' in \emph{Image Processing,
  1998. ICIP 98. Proceedings. 1998 International Conference on}, vol.~1.\hskip
  1em plus 0.5em minus 0.4em\relax IEEE, 1998, pp. 62--66.

\bibitem{Okazawa2015}
G.~Okazawa, S.~Tajima, and H.~Komatsu, ``{Image statistics underlying natural
  texture selectivity of neurons in macaque V4.}'' \emph{Proceedings of the
  National Academy of Sciences of the United States of America}, vol. 112,
  no.~4, pp. E351--60, 2015.

\bibitem{Hiramatsu2011}
C.~Hiramatsu, N.~Goda, and H.~Komatsu, ``{Transformation from image-based to
  perceptual representation of materials along the human ventral visual
  pathway},'' \emph{NeuroImage}, vol.~57, no.~2, pp. 482--494, 2011.

\bibitem{tipping1999}
M.~E. Tipping and C.~Bishop, ``{Probabilistic Principal Component Analysis},''
  \emph{Journal of the Royal Statistical Society, Series B}, vol. 21/3, pp.
  611--622, jan 1999.

\bibitem{Kylberg2011c}
\BIBentryALTinterwordspacing
G.~Kylberg, ``The kylberg texture dataset v. 1.0,'' Centre for Image Analysis,
  Swedish University of Agricultural Sciences and Uppsala University, Uppsala,
  Sweden, External report (Blue series)~35, September 2011. [Online].
  Available: \url{http://www.cb.uu.se/~gustaf/texture/}
\BIBentrySTDinterwordspacing

\bibitem{karakida2016dynamical}
R.~Karakida, M.~Okada, and S.-i. Amari, ``Dynamical analysis of contrastive
  divergence learning: Restricted boltzmann machines with gaussian visible
  units,'' \emph{Neural Networks}, vol.~79, pp. 78--87, 2016.

\end{thebibliography}

\end{document}